\documentclass[conference]{IEEEtran}
\usepackage{cite}
\usepackage{amsmath,amssymb,amsfonts}
\usepackage{algorithmic}
\usepackage{graphicx}
\usepackage{textcomp}
\usepackage{xcolor}
\usepackage{booktabs}  % For professional looking tables (thick/thin lines)
\usepackage{multirow}  % For merging cells vertically

\begin{document}

\title{KD-OCT: Efficient Knowledge Distillation for Clinical-Grade Retinal OCT Classification}

\author{\IEEEauthorblockN{Erfan Nourbakhsh}
\IEEEauthorblockA{Artificial Intelligence Department\\
University of Isfahan\\
Isfahan, Iran\\
erfannourbakhsh2001@gmail.com}
\and
\IEEEauthorblockN{Nasrin Sanjari}
\IEEEauthorblockA{Labbafinejad Hospital\\
Shahid Beheshti University of Medical Science\\
Tehran, Iran\\
nasrinsanjari@mail.mui.ac.ir}
\and
\IEEEauthorblockN{Ali Nourbakhsh}
\IEEEauthorblockA{Department of Mechanical Engineering\\
Isfahan University of Technology\\
Isfahan, Iran\\
nourbakhsh.a@me.iut.ac.ir}}

\maketitle

\begin{abstract}
Age-related macular degeneration (AMD) and choroidal neovascularization (CNV)-related conditions are leading causes of vision loss worldwide, with optical coherence tomography (OCT) serving as a cornerstone for early detection and management. However, deploying state-of-the-art deep learning models like ConvNeXtV2-Large in clinical settings is hindered by their computational demands. Therefore, it is desirable to develop efficient models that maintain high diagnostic performance while enabling real-time deployment. In this study, a novel knowledge distillation framework, termed KD-OCT, is proposed to compress a high-performance ConvNeXtV2-Large teacher model, enhanced with advanced augmentations, stochastic weight averaging, and focal loss, into a lightweight EfficientNet-B2 student for classifying normal, drusen, and CNV cases. KD-OCT employs real-time distillation with a combined loss balancing soft teacher knowledge transfer and hard ground-truth supervision. The effectiveness of the proposed method is evaluated on the Noor Eye Hospital (NEH) dataset using patient-level cross-validation. Experimental results demonstrate that KD-OCT outperforms comparable multi-scale or feature-fusion OCT classifiers in efficiency-accuracy balance, achieving near-teacher performance with substantial reductions in model size and inference time. Despite the compression, the student model exceeds most existing frameworks, facilitating edge deployment for AMD screening.  Code is available at  https://github.com/erfan-nourbakhsh/KD-OCT .
\end{abstract}


\section{Introduction}
Age-related macular degeneration (AMD) is a leading cause of irreversible vision loss globally, representing about 8.7\% of worldwide blindness and mainly impacting those over 60~\cite{wong2014global,taylor2016how}. Designated a priority eye disease by the World Health Organization, its prevalence is expected to surge with aging populations, potentially affecting 288 million people by 2040 \cite{rasti2018macular}. As a chronic disorder, AMD strains healthcare systems and reduces quality of life by causing gradual central vision loss.

AMD manifests in two primary forms: dry and wet. Dry AMD, comprising 80-90\% of cases, is characterized by the accumulation of drusen, extracellular deposits between the retinal pigment epithelium (RPE) and Bruch's membrane, leading to RPE atrophy and photoreceptor loss~\cite{abdelsalam1999drusen,das2019multiscale}. In 10-20\% of instances, dry AMD progresses to wet AMD, involving choroidal neovascularization (CNV), fluid leakage, and rapid retinal damage~\cite{freund1993amd}. Early detection is critical, as treatments like anti-vascular endothelial growth factor (anti-VEGF) injections can mitigate wet AMD progression, though they are costly, require repeated administration, and carry risks of recurrence~\cite{hwang2019aiamd}.

Optical coherence tomography has revolutionized AMD diagnosis as a non-invasive, high-resolution imaging modality that provides cross-sectional views of retinal structures, enabling precise identification of drusen, CNV, and other pathologies~\cite{brezinski2002oct,puliafito1995imaging}. However, manual OCT interpretation is labor-intensive, especially given the volume of scans and the chronic monitoring required for AMD patients. This underscores the need for automated computer-aided diagnosis (CAD) systems to alleviate clinical workloads and improve screening efficiency.

Recent advancements in deep learning have yielded promising OCT classification models, often incorporating multi-scale feature fusion or convolutional neural networks (CNNs) to handle varying lesion sizes~\cite{sotoudehpaima2022multiscale,pang2024novel}. However, state-of-the-art models like ConvNeXtV2-Large~\cite{woo2023convnextv2}, despite high accuracy, remain computationally demanding ($\sim$197M parameters), restricting deployment in resource-limited clinical environments~\cite{hu2018senet}. Knowledge distillation (KD) resolves this by transferring knowledge from large teacher models to compact student models~\cite{hinton2015distilling,li2025kdmedicalimaging}. In KD, the student learns from both hard ground-truth labels and the teacher's softened probability distributions (soft labels), which encode nuanced inter-class relationships and boost generalization. This typically uses a combined loss function balancing cross-entropy on true labels with Kullback-Leibler~\cite{kullback1951information} divergence on teacher-student outputs, enabling efficient compression without significant accuracy loss~\cite{sevinc2025distillation,xu2024ela,hinton2015distilling,li2025kdmedicalimaging,yilmaz2025crossarch}.

In this study, we introduce KD-OCT, a new knowledge distillation framework that compresses a high-performance ConvNeXtV2-Large teacher model—augmented with advanced techniques, stochastic weight averaging, and focal loss—into a compact EfficientNet-B2 student for classifying normal, drusen, and CNV in retinal OCT images. KD-OCT uses real-time distillation via a temperature-scaled combined loss and is assessed on the Noor Eye Hospital (NEH) dataset with patient-level 5-fold cross-validation. Results show KD-OCT attains near-teacher accuracy with 25.5× fewer parameters, surpassing similar multi-scale or feature-fusion OCT classifiers in efficiency-accuracy trade-off, enabling edge deployment for AMD screening.

\section{Related works}
The automated classification of retinal pathologies from OCT images has evolved significantly, driven by the need for efficient screening of AMD and related conditions such as drusen and CNV. Early studies relied on traditional machine learning approaches, which typically involved multi-stage pipelines including preprocessing (e.g., denoising and retinal flattening), manual feature extraction using descriptors like histogram of oriented gradients (HOG), local binary patterns (LBP), or scale-invariant feature transform (SIFT), and classification via algorithms such as support vector machines (SVM) or random forests~\cite{srinivasan2014fully,albarrak2013amd,lemaitre2016lbp}. These methods achieved reasonable results but were limited by the time-consuming nature of feature engineering, expert dependency, and poor generalization across datasets due to variations in interpretations.

With the rise of deep learning (DL), convolutional neural networks (CNNs) have emerged as the foundation for end-to-end OCT classification, automatically extracting hierarchical features without manual input~\cite{ting2019aidl,lecun2015deep}. Classic models like VGG~\cite{simonyan2014vgg}, Inception~\cite{li2021domainadapt}, and ResNet \cite{kumar2024resnet18} have been adapted for retinal disease detection, achieving high accuracy in AMD stage identification~\cite{simonyan2014vgg,li2021domainadapt,kumar2024resnet18}. To tackle varying lesion sizes in OCT images (e.g., small drusen vs. extensive CNV), multi-scale methods have become key. For example, multi-scale deep feature fusion (MDFF) merges features across scales to capture inter-scale differences and boost discriminative ability. Feature pyramid networks (FPN) integrate top-down propagation with lateral connections to retain fine details alongside high-level context, lowering model complexity. Spatial attention mechanisms in multi-scale setups, often with depthwise separable convolutions, highlight pathological areas while managing parameter expansion.

Recently, Transformer-based models have been investigated for their global receptive fields, differing from CNNs' local emphasis\cite{vaswani2017attention,dosovitskiy2020vit}. Vision Transformers (ViT) have been tailored for retinal OCT classification \cite{dosovitskiy2020vit}, including variants like structure-oriented Transformers that integrate clinical priors (e.g., structure-guided modules) for disease grading \cite{shen2023structureoriented}. Hybrid CNN-Transformer models, featuring parallel branches for local and global feature extraction with adaptive fusion, have excelled in multi-class retinal disease tasks \cite{yang2024hrsnet,ma2022hctnet}. ConvNeXt, a Transformer-inspired pure CNN architecture, exhibits robust feature learning on limited data, rendering it ideal as a backbone for OCT analysis \cite{liu2022convnet2020s,hu2018senet}.

State-of-the-art models in medical image analysis, especially for OCT classification, display notable differences in architecture, efficiency, and applicability to AMD detection tasks, as shown in Figure 1. ResNet \cite{he2016resnet} introduced residual learning via skip connections to train very deep CNNs, alleviating vanishing gradients and enabling robust feature extraction in medical imaging, although it depends on local receptive fields and may falter with global dependencies in intricate retinal structures. Conversely, Swin Transformer \cite{liu2021swin} features a hierarchical Vision Transformer with shifted windows for efficient self-attention, grasping multi-scale contextual details and long-range interactions, which excels in dense prediction tasks like OCT segmentation and classification by managing varying lesion scales more adeptly than conventional CNNs. ConvNeXt \cite{liu2022convnet2020s} updates CNNs by adding Transformer-inspired components (e.g., larger kernels, GELU activations) to rival hierarchical Transformers, providing a mix of computational efficiency and performance in resource-limited medical environments. Its successor, ConvNeXtV2 \cite{woo2023convnextv2}, boosts scalability using masked autoencoders for self-supervised pre-training, enhancing representation learning on scarce labeled data common in clinical OCT datasets and delivering superior generalization in multi-class retinal disease tasks over prior versions.

\begin{figure}[h]
    \centering
    \includegraphics[width=\linewidth]{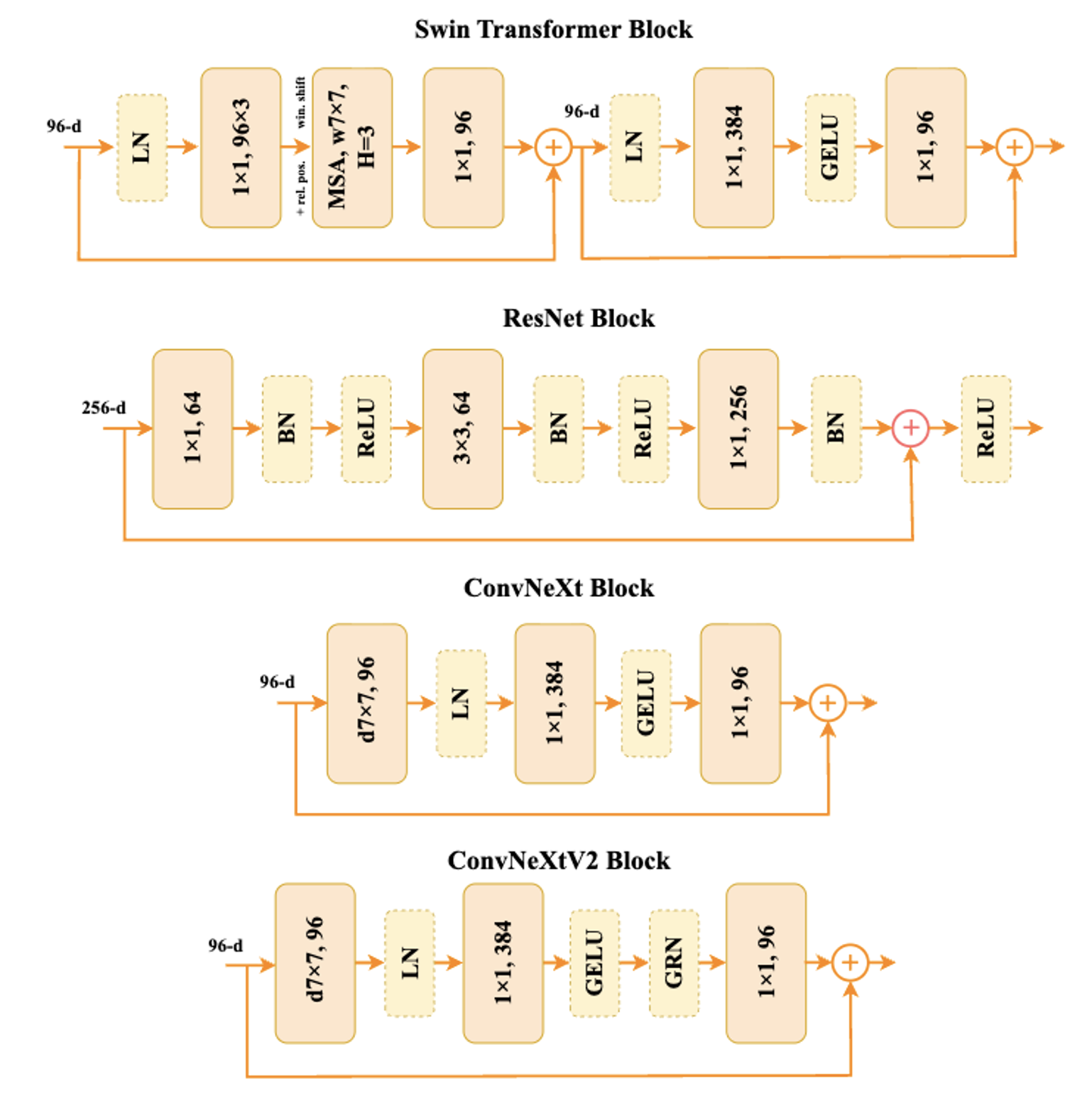}
    \caption{Comparison of block architectures in SOTA models for medical image analysis: (a) Swin Transformer Block \cite{liu2021swin}, featuring shifted window-based multi-head self-attention for efficient hierarchical processing; (b) ResNet Block \cite{he2016resnet}, utilizing residual connections with batch normalization and ReLU activations for deep network training; (c) ConvNeXt Block \cite{liu2022convnet2020s}, incorporating depthwise convolutions, layer normalization, and GELU for Transformer-inspired CNN efficiency; (d) ConvNeXtV2 Block \cite{woo2023convnextv2}, enhancing the prior with global response normalization (GRN) for improved scaling and self-supervised learning.}
    \label{fig:block_architectures}
\end{figure}

While these advancements have boosted accuracy, the computational demands of large models like ConvNeXtV2-Large ($\sim$197M parameters) limit clinical deployment \cite{hu2018senet}. Knowledge distillation (KD) serves as a vital compression method, transferring knowledge from a complex "teacher" to a lightweight "student" through soft labels and intermediate representations \cite{hinton2015distilling}. In medical imaging, KD extends to semi-supervised learning, class balancing, and privacy preservation, as noted in recent surveys \cite{li2025kdmedicalimaging,kullback1951information}. In retinal imaging, multi-task KD enables eye disease prediction from fundus images, with teacher ensembles distilling knowledge across coarse/fine-grained classification and textual diagnosis generation, yielding high performance on limited labeled data \cite{chelaramani2021multitask}. For anomaly detection in retinal fundus images, cross-architecture KD compresses Vision Transformers (ViT) to CNNs for edge deployment on devices like NVIDIA Jetson Nano, maintaining $\sim$93\% of teacher accuracy with 97.4\% fewer parameters \cite{yilmaz2025crossarch}. In OCT-specific applications, fundus-enhanced disease-aware KD transfers unpaired fundus knowledge to OCT models via class prototype matching and similarity alignment, enhancing multi-label retinal disease classification without paired datasets \cite{wang2023fundusenhanced}. Unsupervised anomaly detection in OCT employs Teacher-Student KD, training only on normal scans to detect pathologies (e.g., AMD, DME) and produce anomaly scores and maps for screening \cite{aresta2025anomalydistill}. Equity-enhanced KD has been used for glaucoma progression prediction from OCT, ensuring demographic fairness \cite{afolabi2025equity}.

Despite these advances, gaps remain in applying KD to clinical-grade AMD classification from OCT, particularly in cross-architecture distillation for efficiency, real-time teacher inference to avoid pre-computing labels, and integration with domain-specific enhancements like patient-disjoint validation for robust generalization on imbalanced datasets. Our KD-OCT framework addresses these by compressing an enhanced ConvNeXtV2-Large teacher to an EfficientNet-B2 student, leveraging real-time distillation and tailored augmentations for scalable AMD screening.

\section{Dataset}
The proposed KD-OCT method was evaluated on two publicly available databases to assess its performance in classifying normal, drusen, and CNV cases from retinal OCT images. The primary dataset is the Noor Eye Hospital (NEH) dataset, consisting of anonymized OCT images acquired using the Heidelberg Spectralis SD-OCT imaging system at Noor Eye Hospital, Tehran, Iran \cite{sotoudehpaima2021mendeley}. The images contain no marks, features, or patient identifiers to ensure privacy, and all B-scans were labeled by a retinal specialist. Inclusion criteria included individuals over 50 years of age, absence of any other retinal pathologies, and good image quality (signal strength $Q \ge 20$). To simulate challenging conditions, only the worst-case B-scans per volume were retained (e.g., for CNV patients, scans prominently displaying CNV), resulting in 12,649 B-scans from an original total of 16,822 across 441 patients and 554 eyes. The class distribution includes 5,667 normal scans from 120 patients, 3,742 drusen scans from 160 patients, and 3,240 CNV scans from 161 patients.

The secondary dataset is the University of California San Diego (UCSD) dataset \cite{kermany2018identifying}, which includes four categories: CNV, diabetic macular edema (DME), drusen, and normal. The training set comprises 108,312 retinal OCT images from 4,686 patients, with 37,206 CNV, 11,349 DME, 8,617 drusen, and 51,140 normal images. The test set consists of 1,000 images from 633 patients, evenly distributed with 250 from each category.

\section{Proposed Approach}
\subsection{Data Preparation}
To ensure robust evaluation and fair comparison, the datasets were divided into training, validation, and test sets, as shown in Figure 2. For the Noor Eye Hospital (NEH) dataset, 20\% of the total data was assigned to the test set for independent benchmarking, with the remaining 80\% split into training and validation. From this 80\%, 20\% was set aside for validation to track performance and avoid overfitting, leaving the rest for training. This stratified split occurred at the patient level to prevent data leakage, ensuring no patient scan overlap across sets and enhancing generalization in clinical settings. For the UCSD dataset, the predefined test set of 1,000 images was kept unchanged, while the 108,312-image training set was subdivided with 20\% for validation and the remainder for training. This setup aligns with baseline methods, like the Multi-Scale Convolutional Neural Network \cite{sotoudehpaima2022multiscale}, which used comparable validation ratios to tune hyperparameters and assess performance on imbalanced retinal OCT classes.

\begin{figure}[h]
    \centering
    \includegraphics[width=\linewidth]{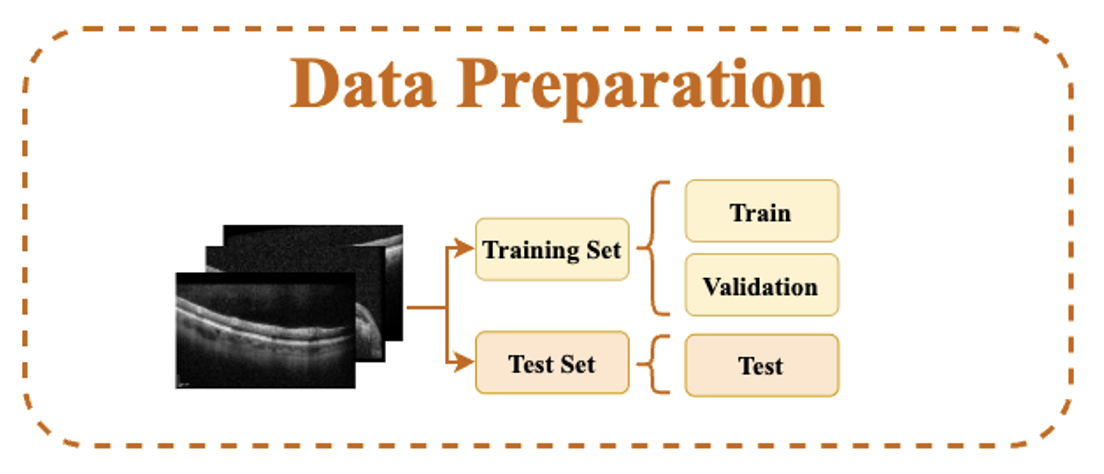}
    \caption{Overview of data preparation.}
    \label{fig:data_preparation}
\end{figure}

\subsection{Data Augmentation}

Data augmentation is vital in the KD-OCT framework, artificially enlarging the training dataset, boosting model robustness, and reducing overfitting, especially in knowledge distillation, where the student gains from varied inputs to replicate the teacher's generalizations on imbalanced retinal OCT data. As shown in Figure 3, the augmentation approach is customized for training, validation, and inference phases to balance complexity and efficiency while maintaining clinical relevance, including managing variations in scan orientation, lighting, and artifacts typical in OCT imaging.

For the training pipeline, a comprehensive sequence of transformations is applied to introduce variability and simulate real-world imperfections in retinal scans. The process begins with resizing the image to a larger square dimension, followed by a random crop to a target square size, which normalizes dimensions while introducing spatial diversity to focus on varying retinal regions. We then apply a fixed number of random operations from a set including brightness, contrast, saturation, sharpness, rotation, and translation adjustments, automating policy selection to improve generalization without manual tuning. Subsequent steps include rotations to simulate probe orientation differences, affine transformations with shear and scale parameters for geometric distortions like misalignments due to patient movement, and color adjustments with brightness, contrast, saturation, and hue factors to account for intensity variations across devices. Horizontal and vertical flips, each with a specified probability, add symmetry invariance, mimicking left/right or top/bottom scan flips. Blurring with a kernel size and probability emulates blurry or noisy acquisitions, while bit reduction with probability handles quantization effects from compression. The image is then converted to a normalized tensor range, followed by erasing with probability and scale range to simulate occlusions like blood vessels or artifacts, and finally normalized using ImageNet-derived mean and standard deviation statistics for consistency with pretrained models. The output is a normalized tensor in channel-height-width format, promoting resilience to clinical variabilities in OCT scans.

The validation pipeline is kept minimal to evaluate the model on near-original data, consisting of resizing to a target square dimension, conversion to a normalized tensor range, and normalization with the same statistics as training. This ensures an unbiased assessment without introducing training-like variability.

For inference, Test-Time Augmentation (TTA), a technique that applies data augmentations during inference to generate multiple input variants and ensembles their predictions for improved reliability and reduced uncertainty \cite{wang2019aleatoric}, is employed to boost prediction reliability by generating multiple augmented versions of each input and averaging their outputs. The five augmentations include: (1) the original resized and normalized image; (2) horizontal flip after resize and normalize; (3) vertical flip after resize and normalize; (4) resize to a larger dimension followed by center crop to the target size and normalize; and (5) resize with small rotation and normalize. This produces a list of five normalized tensors, whose averaged logits enhance accuracy and robustness, particularly for subtle AMD features in OCT, by reducing sensitivity to minor input perturbations without additional training overhead.

\begin{figure}[h]
    \centering
    \includegraphics[width=\linewidth]{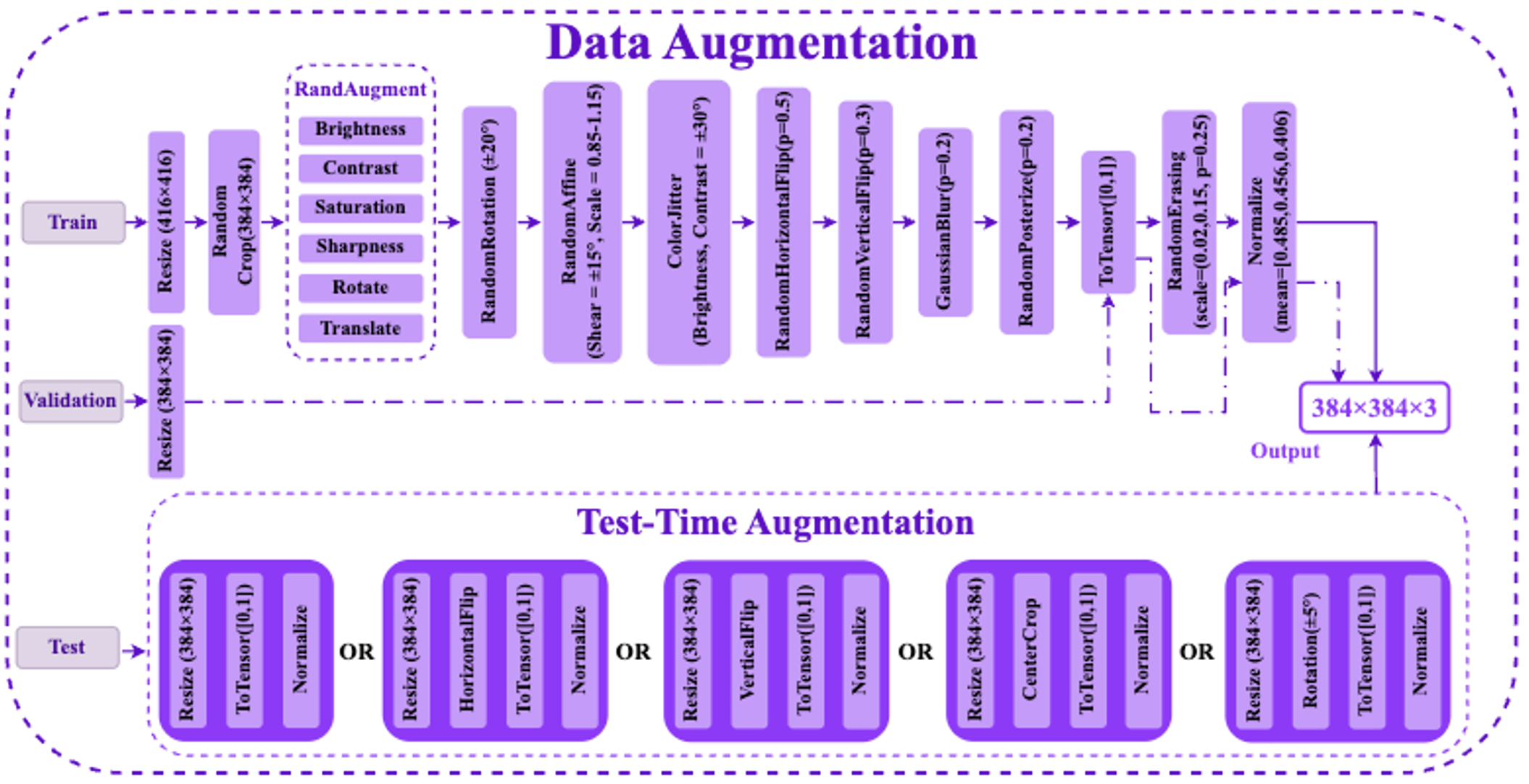}
    \caption{Overview of the data augmentation pipelines in KD-OCT, including the training sequence with RandAugment and geometric/color transforms, minimal validation steps, and Test-Time Augmentation (TTA) variants for inference.}
    \label{fig:augmentation_pipeline}
\end{figure}

\subsection{Teacher Model Architecture}
The core of the KD-OCT teacher model uses the ConvNeXtV2-Large backbone \cite{woo2023convnextv2}, a cutting-edge convolutional neural network that integrates Transformer-inspired efficiencies while preserving CNN strengths in inductive biases and computational scalability. Pretrained on ImageNet-22K and fine-tuned on ImageNet-1K via Fully Convolutional Masked AutoEncoder (FCMAE) \cite{woo2023convnextv2} for self-supervised learning, it features a large parameter count and handles input images in batch-channel-height-width format. A drop path rate provides stochastic depth regularization to boost generalization during training. As shown in Figure 4, the architecture includes a stem layer, four hierarchical stages with downsampling transitions, and a classification head, supporting progressive feature extraction from low-level details to high-level semantics for classifying retinal OCT scans as normal, drusen, or CNV.

The stem layer initializes feature processing with a convolutional kernel and stride, expanding input channels, followed by LayerNorm, resulting in an output with increased channels and reduced spatial dimensions. Stage 1 focuses on early feature extraction with several ConvNeXtV2 blocks at initial channels and resolution (with progressive drop path), each comprising DepthWise Conv, LayerNorm, Linear expansion, GELU activation, Global Response Normalization (GRN), and Linear reduction back to base channels, yielding the same dimension. Downsampling to Stage 2 uses LayerNorm and convolutional stride, doubling channels and halving resolution. Stage 2 employs blocks with similar components but expanded intermediate channels, outputting the updated dimension.

\begin{figure}[h]
    \centering
    \includegraphics[width=\linewidth]{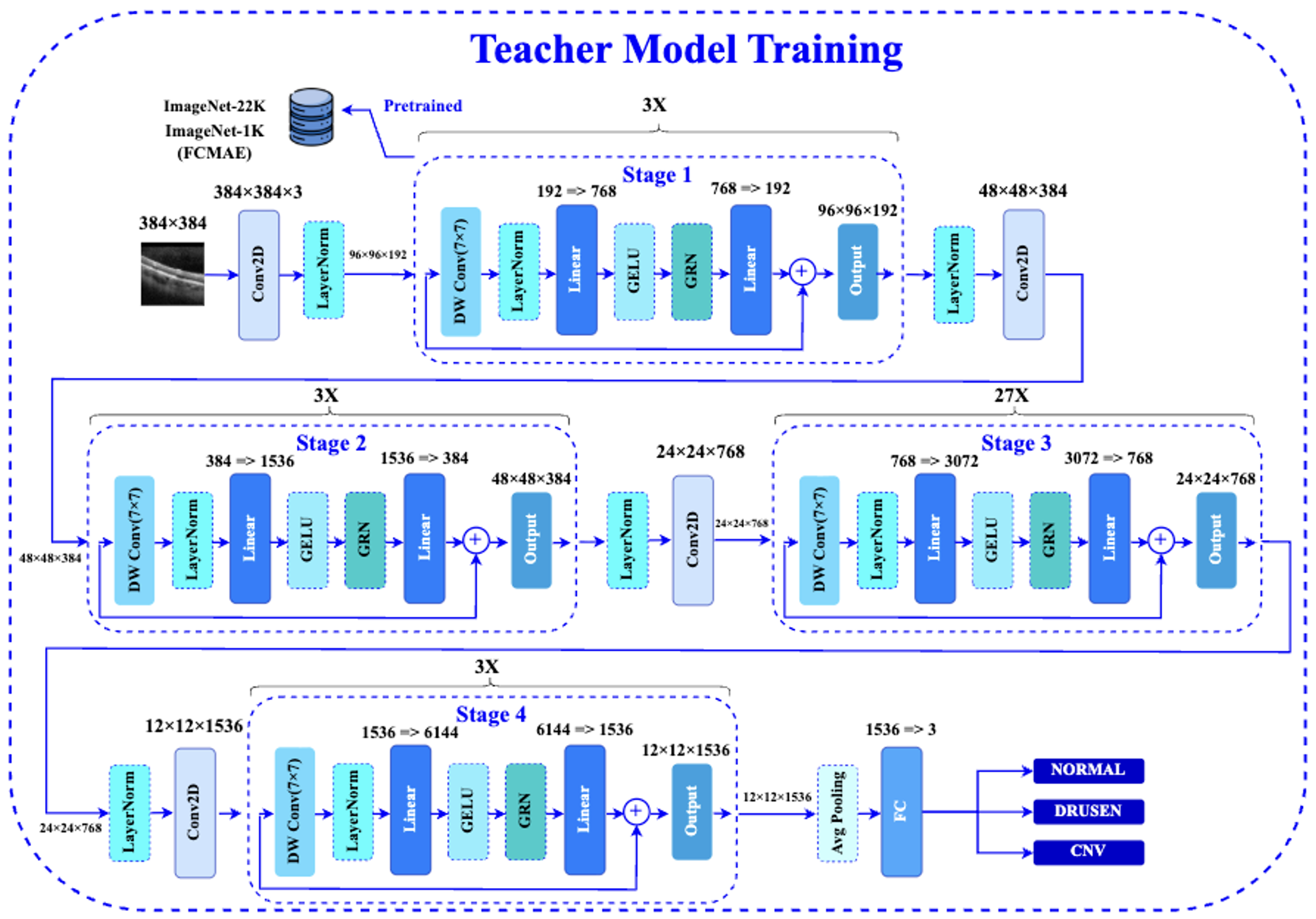}
    \caption{Overview of the teacher model training.}
    \label{fig:teacher_training}
\end{figure}

Further downsampling to Stage 3 (LayerNorm + convolutional stride) increases channels while reducing resolution. This primary feature extraction stage, the deepest with numerous blocks (progressive drop path) and substantial intermediate expansion, captures intricate retinal patterns like drusen deposits or CNV membranes, producing the stage output. The final downsampling to Stage 4 yields even higher channels at smaller resolution, processed by blocks with large expansion, outputting the final backbone features. The classification head applies global average pooling to reduce spatial dimensions, followed by dropout for regularization, and a fully connected layer to generate raw logits for multi-class prediction without additional activation.

\subsection{Knowledge Distillation}
Integrating the preceding components, data preparation, augmentation, and teacher model architecture, the KD-OCT framework employs knowledge distillation to transfer expertise from the high-capacity ConvNeXtV2-Large teacher to the lightweight EfficientNet-B2 student \cite{lin2020focal}, enabling efficient deployment while preserving clinical-grade performance in retinal OCT classification, as illustrated in Figure 5. This cross-architecture distillation process \cite{yilmaz2025crossarch} first involves training the teacher model end-to-end on the prepared and augmented data using focal loss \cite{lin2020focal} to handle class imbalance, stochastic weight averaging (SWA) for smoother convergence, and advanced techniques like differential learning rates (head: 1e-4, backbone: 2e-5) with AdamW optimization \cite{loshchilov2017adamw}, weight decay to prevent overfitting by regularizing model weights, 10-epoch warmup, and cosine annealing scheduler \cite{loshchilov2017sgdr} over up to 150 epochs. The teacher's heavy augmentation pipeline ensures robust feature learning, capturing nuanced retinal pathologies like subtle drusen or CNV irregularities.

The focal loss is defined as:

\begin{equation}
    \mathcal{FL} = \alpha_t \cdot (1 - \rho_t)^{\gamma} \cdot \log(\rho_t)
    \label{eq:focal_loss}
\end{equation}

where $\alpha_t$ is the class weighting factor, $\rho_t$ is the predicted probability for the true class, and $\gamma$ is the focusing parameter (typically set to 2.0 in our experiments) that down-weights easy examples to emphasize hard ones.

The cosine annealing scheduler adjusts the learning rate as:

\begin{equation}
    lr = min\_lr + (base\_lr - min\_lr) \times 0.5 \times (1 + \cos(\pi \cdot progress))
    \label{eq:cosine_annealing}
\end{equation}

where $base\_lr$ is the initial learning rate, $min\_lr$ is the minimum learning rate, and $progress$ is the fractional progress through the annealing cycle.

\begin{figure}[h]
    \centering
    \includegraphics[width=\linewidth]{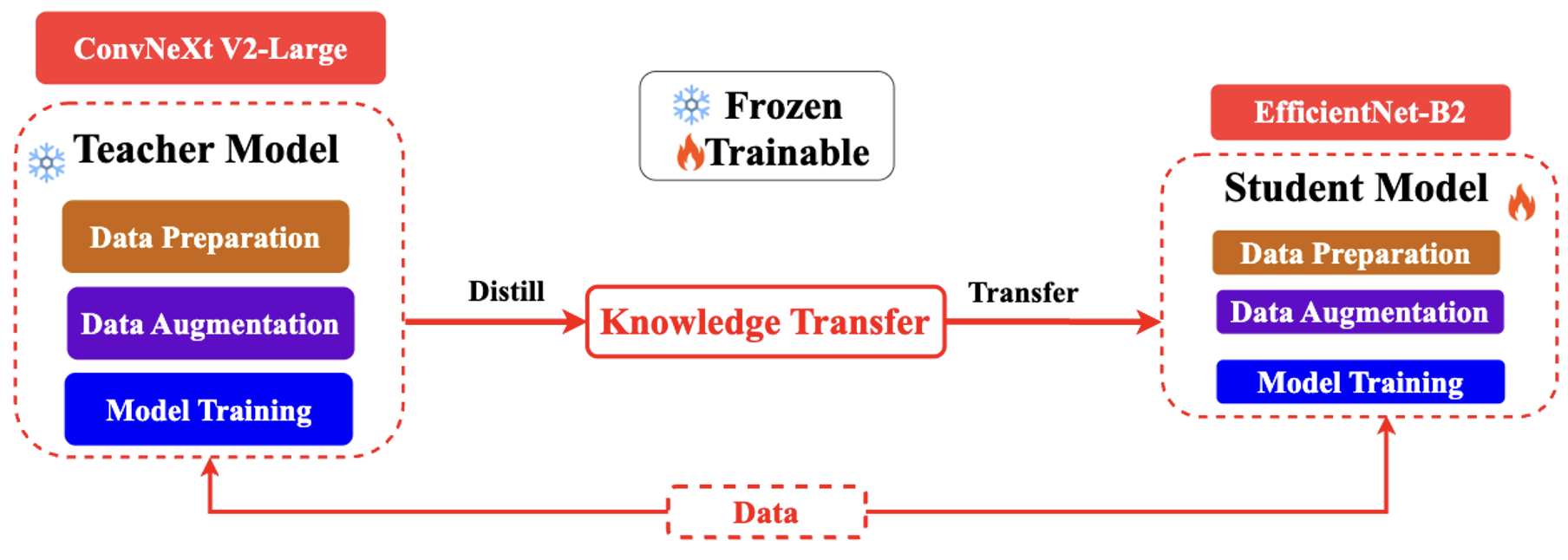}
    \caption{Overview of the KD-OCT framework, showing knowledge transfer from the ConvNeXtV2-Large teacher to the EfficientNet-B2 student via real-time distillation.}
    \label{fig:kdoct_framework}
\end{figure}

Once trained, real-time KD is performed where the frozen teacher generates soft labels on-the-fly during student training, avoiding offline logit pre-computation and allowing dynamic knowledge transfer adapted to the student's progress \cite{li2025kdmedicalimaging}. The student, based on EfficientNet-B2, uses a lighter augmentation strategy (e.g., reduced number of random operations to limit intensity, milder rotations to simulate subtle variations without excessive distortion, no blur/posterize) and a unified learning rate with AdamW (weight decay to prevent overfitting by regularizing model weights, warmup period to stabilize initial training, cosine annealing scheduler to gradually reduce the learning rate for better convergence over multiple epochs, early stopping patience to halt training when validation performance plateaus). The combined loss balances low-weighted cross-entropy for hard ground-truth labels with high-weighted, temperature-scaled Kullback-Leibler divergence for soft teacher knowledge, helping the student learn inter-class similarities and generalize on imbalanced datasets without focal loss or SWA. Batch configurations maintain an effective size (teacher: smaller batch size with higher accumulation steps; student: larger batch size with fewer accumulation steps) using FP16 mixed precision for efficiency. This approach compresses the model for edge deployment and outperforms baselines in efficiency-accuracy trade-offs for AMD screening.

\section{Hyper-parameters}
The KD-OCT framework uses finely tuned hyperparameters to optimize performance and enable efficient knowledge transfer from the ConvNeXtV2-Large teacher to the EfficientNet-B2 student. Key configurations include differential learning rates for the teacher ($1\times10^{-4}$ for classification head, $2\times10^{-5}$ for backbone) with 0.05 weight decay, 10-epoch linear warmup, and cosine annealing scheduler decaying to $1\times10^{-7}$ over up to 150 epochs (early stopping patience 25), while the student employs a unified $1\times10^{-3}$ learning rate, 0.01 weight decay, 5-epoch warmup, and cosine annealing to $1\times10^{-6}$ over a maximum of 100 epochs (patience 20). Both leverage AdamW optimization and FP16 mixed precision training with an effective batch size of 16 via gradient accumulation (teacher: batch size 4, accumulation 4; student: batch size 8, accumulation 2). Distillation applies a 4.0 temperature for soft labels, with loss weights balancing hard supervision ($\beta=0.3$, cross-entropy) and soft transfer ($\alpha=0.7$, Kullback-Leibler divergence). Augmentations feature RandAugment ($N=2$, $M=9$ for teacher; $M=7$ for student), rotations ($\pm20^{\circ}$ teacher; $\pm15^{\circ}$ student), and TTA using 5 variants to boost robustness. Training occurred on an NVIDIA H200 GPU, utilizing its high memory bandwidth to manage large models and batches effectively.

\begin{table*}[h]
\centering
\caption{The Results of a three-class classification task on the NEH dataset, evaluated using five-fold cross-validation. (*The results of this model are directly reported from\cite{sotoudehpaima2022multiscale}.)}
\label{tab:results}
\renewcommand{\arraystretch}{1.2}
\begin{tabular}{lcccc}
\hline
\textbf{Models} & \textbf{Param (mil)} & \textbf{Accuracy} & \textbf{Sensitivity} & \textbf{Specificity} \\
\hline
HOG + SVM* & -- & $67.2 \pm 3.7$ & $66.99 \pm 3.1$ & $74.3 \pm 2.5$ \\
\hline
VGG16* \cite{simonyan2014vgg} & 28.3 & $91.6 \pm 2.2$ & $91.4 \pm 2.0$ & $95.6 \pm 1.1$ \\
ResNet50* \cite{he2016resnet} & 23.6 & $86.8 \pm 2.0$ & $86.4 \pm 1.6$ & $93.0 \pm 0.9$ \\
DenseNet121* \cite{huang2017densenet} & 7.0 & $90.0 \pm 1.4$ & $89.7 \pm 1.7$ & $94.7 \pm 0.8$ \\
EfficientNetB0* \cite{tan2019efficientnet} & 4.0 & $85.4 \pm 2.6$ & $84.5 \pm 2.2$ & $92.1 \pm 1.3$ \\
Kermany et al.* \cite{kermany2018identifying} & 0.02 & $83.9 \pm 1.7$ & $82.9 \pm 2.3$ & $91.4 \pm 1.0$ \\
Kaymak et al.* \cite{kaymak2018amd_dme} & 58.3 & $80.2 \pm 4.7$ & $80.0 \pm 4.4$ & $89.4 \pm 2.5$ \\
Thomas et al.* \cite{thomas2021multiscale} & 2.5 & $68.5 \pm 4.9$ & $69.1 \pm 4.3$ & $83.8 \pm 2.8$ \\
FPN-VGG16* \cite{sotoudehpaima2022multiscale} & 21.6 & $92.0 \pm 1.6$ & $91.8 \pm 1.7$ & $95.8 \pm 0.9$ \\
FPN-ResNet50* \cite{sotoudehpaima2022multiscale} & 31.1 & $90.1 \pm 2.9$ & $89.8 \pm 2.8$ & $94.8 \pm 1.4$ \\
FPN-DenseNet121* \cite{sotoudehpaima2022multiscale} & 14.3 & $90.9 \pm 1.4$ & $90.5 \pm 1.9$ & $95.2 \pm 0.7$ \\
FPN-EfficientNetB0* \cite{sotoudehpaima2022multiscale} & 12.7 & $87.8 \pm 1.3$ & $86.6 \pm 1.8$ & $93.3 \pm 0.8$ \\
SF net \cite{zheng2025sfnet} & 29.2 & $82.6 \pm 2.4$ & $80.4 \pm 2.8$ & $96.2 \pm 0.6$ \\
MedSigLIP \cite{sellergren2025medgemma} & 430.4 & $84.5 \pm 3.2$ & $81.81 \pm 4.64$ & $94.42 \pm 1.09$ \\
\hline
KD-OCT (Ours) -- ConvNeXtV2-Large & 196.4 & $\mathbf{92.6 \pm 2.3}$ & $\mathbf{92.9 \pm 2.1}$ & $\mathbf{98.1 \pm 0.8}$ \\
KD-OCT (Ours) -- EfficientNet-B2 & \textbf{7.7} & $\mathbf{92.46 \pm 1.36}$ & $\mathbf{92.15 \pm 1.29}$ & $\mathbf{96.04 \pm 0.78}$ \\
\hline
\end{tabular}
\end{table*}

\section{Results}
The experimental results demonstrate the KD-OCT framework's superior efficacy in retinal OCT classification, balancing high accuracy with computational efficiency for clinical deployment. On the NEH dataset, evaluated via five-fold patient-level cross-validation for three-class classification (normal, drusen, CNV; Table I), the ConvNeXtV2-Large teacher achieved 92.6\% accuracy, outperforming baselines such as FPN-VGG16 (92.0\%) \cite{sotoudehpaima2022multiscale} and MedSigLIP (84.5\%) \cite{sellergren2025medgemma}. This highlights the teacher's advanced architecture and robust training, including focal loss and heavy augmentations, for handling class imbalances and subtle pathologies like early drusen or CNV.

Even more compelling is the performance of the distilled EfficientNet-B2 student model on the same NEH dataset, attaining 92.46\% accuracy, nearly matching the teacher, while drastically reducing model size from 196.4 million to just 7.7 million parameters, a 25.5× compression. This not only surpasses multi-scale competitors like FPN-DenseNet121 (90.9\% accuracy) \cite{sotoudehpaima2022multiscale} and SF Net (82.6\% accuracy) \cite{zheng2025sfnet} but also underscores KD-OCT's strength in knowledge transfer, where the student inherits the teacher's nuanced understanding without the computational overhead, making it ideal for resource-limited clinical settings like portable OCT devices.

\begin{table}[h]
\centering
\caption{Results of a four-class classification task on the UCSD dataset. (* The results of this model are directly reported from\cite{sotoudehpaima2022multiscale}.)}
\label{tab:ucsd_results}
\renewcommand{\arraystretch}{1.2}
\resizebox{\columnwidth}{!}{%
\begin{tabular}{lcccc}
\hline
\textbf{Models} & \textbf{Preprocess} & \textbf{Accuracy} & \textbf{Sensitivity} & \textbf{Specificity} \\
\hline
VGG16* \cite{simonyan2014vgg}          & \textcolor{red}{$\times$} & 93.9 & 100   & 90.8  \\
ResNet50* \cite{he2016resnet}       & \textcolor{red}{$\times$} & 96.7 & 99.6  & 94.8  \\
EfficientNetB0* \cite{tan2019efficientnet} & \textcolor{red}{$\times$} & 95.0 & 99.8  & 91.4  \\
Kermany et al.* \cite{kermany2018identifying} & \textcolor{red}{$\times$} & 96.6 & 97.8  & 97.4  \\
Kaymak et al.* \cite{kaymak2018amd_dme}  & \textcolor{red}{$\times$} & 97.1 & 98.4  & 99.6  \\
Hassan et al.* \cite{hassan2020ragfw}  & \textcolor{green!60!black}{$\checkmark$} & 98.6 & 98.27 & 99.6  \\
FPN-VGG16* \cite{sotoudehpaima2022multiscale}      & \textcolor{red}{$\times$} & 98.4 & \textbf{100}   & 97.4  \\
\hline
KD-OCT (Ours) ConvNeXtV2-Large & \textcolor{red}{$\times$} & 98.4 & 98.45 & 99.47 \\
KD-OCT (Ours) EfficientNet-B2  & \textcolor{red}{$\times$} & \textbf{98.4} & 98.40 & \textbf{99.47} \\
\hline
\end{tabular}}
\end{table} 

To validate generalizability, KD-OCT was tested on the UCSD dataset for four-class classification (normal, drusen, CNV, DME) using the predefined test set (Table II). Without fine-tuning or domain adaptation, both teacher and student models achieved 98.4\% accuracy, outperforming baselines like Hassan et al. (98.6\%, but requiring preprocessing) \cite{hassan2020ragfw} and FPN-VGG16 (98.4\%) \cite{sotoudehpaima2022multiscale}. This seamless transfer across datasets, despite imaging system differences and an added DME class, illustrates the framework's robustness, as distilled knowledge enables high-fidelity predictions on unseen data from diverse clinical environments.

In a more stringent five-fold cross-validation on the UCSD training set (Table III), the teacher and student models further excelled with accuracies of 97.72\% and 97.74\%, respectively, eclipsing multi-scale approaches like Fang et al. (TMI) (90.1\% accuracy) \cite{fang2019attention} and FPN-VGG16 (93.9\% accuracy) \cite{sotoudehpaima2022multiscale}. These consistent gains highlight KD-OCT's generalization advantage, with cross-architecture distillation preserving diagnostic precision while reducing inference time, enabling scalable real-time AMD screening globally.

\begin{table}[h]
\centering
\caption{Results of a four-class classification task on the UCSD dataset, evaluated using five-fold cross-validation (* The results of this model are directly reported from\cite{sotoudehpaima2022multiscale}.)}
\label{tab:ucsd_results}
\renewcommand{\arraystretch}{1.3}
\resizebox{\columnwidth}{!}{%
\begin{tabular}{lcccc}
\hline
\textbf{Models} & \textbf{Preprocess} & \textbf{Accuracy} & \textbf{Sensitivity} & \textbf{Specificity} \\
\hline
Fang et al. (JVCIR)* \cite{fang2019iterative}  & \textcolor{red}{$\times$} & 87.3 & 84.7 & 95.8 \\
Fang et al. (TMI)* \cite{fang2019attention}   & \textcolor{green!60!black}{$\checkmark$} & 90.1 & 86.6 & 96.6 \\
FPN-VGG16* \cite{sotoudehpaima2022multiscale}            & \textcolor{red}{$\times$} & 93.9 & 93.4 & 98.0 \\
\hline
KD-OCT (Ours) ConvNeXtV2-Large & \textcolor{red}{$\times$} & 97.72 & 97.72 & \textbf{99.26} \\
KD-OCT (Ours) EfficientNet-B2  & \textcolor{red}{$\times$} & \textbf{97.74} & \textbf{97.74} & 99.21 \\
\hline
\end{tabular}}
\end{table}

To further elucidate the contributions of the key enhancements in the teacher model, an ablation study was conducted on the NEH dataset using five-fold cross-validation. Removing advanced augmentations (replacing with basic resizing and normalization) reduced the teacher's accuracy, sensitivity, and specificity, emphasizing their role in enhancing robustness to clinical variabilities like scan orientation and artifacts. Excluding stochastic weight averaging caused a moderate performance decline, as it supports smoother optimization and better generalization on imbalanced classes. Omitting focal loss (reverting to standard cross-entropy) led to the largest drop, highlighting its value in tackling class imbalance by focusing on hard examples such as subtle CNV cases. Collectively, these enhancements improved the student's distilled performance over a baseline, preserving near-teacher quality for efficient deployment.

\section{Conclusion and future works}
In this study, we introduced KD-OCT, a novel knowledge distillation framework that compresses a high-performance ConvNeXtV2-Large teacher model—enhanced with advanced augmentations, focal loss, and stochastic weight averaging—into a lightweight EfficientNet-B2 student for classifying normal, drusen, and CNV in retinal OCT images. Using real-time distillation with a temperature-scaled combined loss (balancing soft teacher knowledge and hard ground-truth supervision), KD-OCT attains near-teacher accuracy ($\sim$92-98\%) with 25.5× parameter reduction and faster inference, surpassing multi-scale and feature-fusion baselines in efficiency-accuracy trade-off on the NEH and UCSD datasets. This cross-architecture method, with patient-disjoint cross-validation and tailored augmentations, overcomes computational barriers in clinics, promoting robust generalization on imbalanced classes and edge deployment for scalable AMD screening. Future work will explore semi-supervised KD to reduce labeled data reliance, multi-modal distillation with fundus images for improved accuracy, and extension to other retinal pathologies like diabetic macular edema, while optimizing for real-time integration in portable devices.

\bibliographystyle{IEEEtran}
\bibliography{reference}

\end{document}